\documentclass[journal]{IEEEtran}

\usepackage{microtype}
\usepackage{graphicx}
\usepackage{booktabs} 
\usepackage{xcolor}
\usepackage{subcaption}
\usepackage{hyperref}
\usepackage{multirow}

\usepackage{amsthm}
\usepackage{amsfonts}
\usepackage{thmtools}
\usepackage{thm-restate}
\usepackage{mathtools}

\declaretheorem[name=Definition]{definition}

\newcommand \TODO [1] {}

\newcommand \mc \mathcal

\newcommand {\augm} [1] { \pmb{{#1}}}  

\newcommand {\x} {\augm x}

\newcommand{\cpol}[0]{\mu}

\newcommand {\hide} [1] {}

\ifCLASSINFOpdf
\else
\fi
\usepackage{algorithm}
\usepackage{algorithmic}
\begin{document}
%
\title{Delay-Aware Multi-Agent Reinforcement Learning for Cooperative and Competitive Environments}
%
%
%


\author{Baiming~Chen, Mengdi~Xu, Zuxin~Liu, Liang~Li, Ding~Zhao%
\thanks{Baiming Chen and Liang Li is with the State Key Laboratory of Automotive Safety and Energy, Tsinghua University, Beijing 100084, China.}
\thanks{Mengdi Xu, Zuxin Liu and Ding Zhao are with the Department of Mechanical Engineering, Carnegie Mellon University, Pittsburgh, PA 15213, USA.}
}
\maketitle

\begin{abstract}
Action and observation delays exist prevalently in the real-world cyber-physical systems which may pose challenges in reinforcement learning design. It is particularly an arduous task when handling multi-agent systems where the delay of one agent could spread to other agents. To resolve this problem, this paper proposes a novel framework to deal with delays as well as the non-stationary training issue of multi-agent tasks with model-free deep reinforcement learning. We formally define the Delay-Aware Markov Game that incorporates the delays of all agents in the environment. To solve Delay-Aware Markov Games, we apply centralized training and decentralized execution that allows agents to use extra information to ease the non-stationarity issue of the multi-agent systems during training, without the need of a centralized controller during execution. Experiments are conducted in multi-agent particle environments including cooperative communication, cooperative navigation, and competitive experiments. We also test the proposed algorithm in traffic scenarios that require coordination of all autonomous vehicles to show the practical value of delay-awareness. Results show that the proposed delay-aware multi-agent reinforcement learning algorithm greatly alleviates the performance degradation introduced by delay. Codes and demo videos are available at: https://github.com/baimingc/delay-aware-MARL.
\end{abstract}


%
\IEEEpeerreviewmaketitle

%
\IEEEpeerreviewmaketitle

\section{Introduction}
%
%
%
%

\IEEEPARstart{D}{eep} reinforcement learning (DRL) has made rapid progress in solving challenging problems~\cite{mnih2013playing, silver2016mastering}. Recently, DRL has been used in multi-agent scenarios since many important applications involve multiple agents cooperating or competing with each other, including multi-robot control \cite{matignon2012coordinated}, the emergence of multi-agent communication and language \cite{foerster2016learning, mordatch2018emergence, sukhbaatar2016learning}, multi-player games \cite{peng2017multiagent}, etc. Learning in multi-agent scenarios is fundamentally more difficult than the single-agent case due to many reasons, e.g., non-stationarity \cite{hernandez2017survey}, curse of dimensionality \cite{bu2008comprehensive}, multi-agent credit assignment \cite{agogino2004unifying}, global exploration \cite{matignon2012independent}. For a more comprehensive review of DRL applied in multi-agent scenarios, readers are referred to \cite{hernandez2019survey}.


Most DRL algorithms are designed for synchronous systems with instantaneous observation and action actuation. However, they are not able to handle the delay problem which is prevalent in many real-world applications such as robotic systems~\cite{lazarevic2006finite} and parallel computing \cite{hannah2018unbounded}.
This issue is even worse in multi-agent scenarios, where the delay of one agent could spread to other coupled agents. For example, in tasks involving communications between agents, the action delay of a speaker would give rise to observation delays of all listeners subscribing to this speaker. Ignoring agent delays would not only lead to performance degradation of the agents but also destabilize dynamic systems \cite{gu2003survey}, which may cause catastrophic failures in safety-critical systems. One typical example is the connected and autonomous vehicles~\cite{gong2016constrained}. There are many sources of delays in an autonomous driving system, such as communication delay, sensor delay, time for decision making and actuator delay of the powertrain and the hydraulic brake system. The total delay time could add up to seconds~\cite{bayan2009brake,rajamani2011vehicle,wang2018delay}, which must be properly handled for both performance and safety of the connected vehicle systems.

The time-delay problem has been long studied by the control community. Several methods have been proposed such as Artstein reduction~\cite{artstein1982linear, moulay2008finite}, Smith predictor~\cite{astrom1994new, matausek1999modified} and $H_{\infty}$ controller \cite{mirkin2000extraction}. However, most of these methods assume a known model or make heavy assumptions~\cite{niculescu2001delay, gu2003survey}, which is usually not realistic for real-world tasks. 

Recently, DRL has provided a promising way to solve complex sequential decision making tasks without the assumption of a known model. By modeling the problem as a Markov Decision Processes (MDP), DRL aims to find the optimal policy for the MDP~\cite{sutton2018reinforcement}. However, in time-delayed systems, ignoring the delay violates the Markov property and may lead to arbitrarily suboptimal policies~\cite{singh1994learning}.
To retrieve the Markov property, Walsh \textit{et al.}~\cite{walsh2009learning} reformulated the MDP by augmenting the state space.
Firoiu \textit{et al.}~\cite{firoiu2018human} later used the augmented MDP to solve Atari games with action delay.
Ramstedt \& Pal \cite{ramstedt2019real} proposed a model-free DRL algorithm to efficiently solve the problem with action delays. 
The delay issue could also be addressed with the model-based manner by learning a dynamics model to predict the future state as in \cite{walsh2009learning, firoiu2018human,chen2020delayaware}. However, dealing with multi-agent tasks using model-based DRL involves \textit{agents modeling agents} which introduces extra non-stationarity issues since policies of all agents are consistently updated \cite{hernandez2019survey}.

Most of the previous works are limited to single-agent tasks and are not able to directly handle the non-stationarity issue introduced by multiple agents. 
In this paper, we propose a novel framework to deal with delays as well as the non-stationarity training issue of multi-agent tasks with model-free DRL. The contribution of this paper is three-fold:
\begin{itemize}
    \item We formulate a general model for multi-agent delayed systems, Delay-Aware Markov Game (DA-MG), by augmenting standard Markov Game with agent delays. We prove the solidity of this new structure with the Markov reward process.
    \item We develop a delay-aware training algorithm for DA-MGs that utilizes centralized training and decentralized execution to stabilize the multi-agent training.
    \item We test our algorithm in both benchmark platforms and practical traffic scenarios. 
\end{itemize}




The rest of the paper is organized as follows. We first review the preliminaries in Section~\ref{sec:rel}. In Section~\ref{sec:damg}, we formally define the Delay-Aware Markov Game (DA-MGs) and prove the solidity of this new structure with the Markov reward process. In Section~\ref{sec:damaddpg}, we introduce the proposed framework of delay-aware multi-agent reinforcement learning for DA-MGs with centralized training and decentralized execution. In Section~\ref{sec:experiment}, we demonstrate the performance of the proposed algorithm in cooperative and competitive multi-agent particle environments, as well as traffic scenarios that require coordination of autonomous vehicles.

\section{Preliminaries}
\label{sec:rel}

\subsection{Markov Decision Process and Markov Game}

In the framework of reinforcement learning, the problem is often represented by a Markov Decision Process (MDP). The definition of a standard delay-free MDP is:

\begin{definition}
\label{def:MDP}
A Markov Decision Process (MDP) is characterized by a tuple with \\
(1) state space $\mathcal{S}$, \hspace{0.15cm} (2) action space $\mathcal{A}$, \hspace{0.15cm} \\
(3) initial state distribution $\rho: \mathcal{S} \to \mathbb R$,\\
(4) transition distribution {$p: \mathcal{S} \times \mathcal{A} \times \mathcal{S} \to \mathbb R$}, \hspace{0.15cm}\\
(5) reward function $r: \mathcal{S} \times \mathcal{A} \to \mathbb R$.
\end{definition}


The agent is represented by a policy $\pi$ that directs the action selection, given the current observation.
The goal of the agent is to find the optimal policy $\pi^*$ that maximizes its expected return $G = \Sigma_{t=0}^{T}\gamma^tr\left(s_t, a_t\right)$ where $\gamma$ is a discount factor and $T$ denotes the time horizon.

Markov game is a multi-agent extension of MDP with partially observable environments.
The definition of a standard delay-free Markov game is:

\begin{definition}
\label{def:MG}
A Markov Game (MG) for $N$ agents is characterized by a tuple with \\
(1) A state space $\mathcal{S}$ describing all agents, \\
(2) A set of action spaces $\mathcal{A} = \{\mathcal{A}_1, \dots,\mathcal{A}_N \}$, \\
(3) A set of observation spaces $\mathcal{O} = \{\mathcal{O}_1, \dots,\mathcal{O}_N \}$, \\
(4) initial state distribution $\rho: \mathcal{S} \to \mathbb R$,\\
(5) transition distribution {$p: \mathcal{S} \times \mathcal{A} \times \mathcal{S} \to \mathbb R$},\\
(6) reward function $r_i: \mathcal{S} \times \mathcal{A}_i \to \mathbb R$ for each agent $i$.
\end{definition}

Each agent $i$ receives an individual observation from the state $o_i : \mathcal{S} \to \mathcal{O}_i$ and uses a policy $\pi_i: \mathcal{O}_i \times \mathcal{A}_i \to \mathbb R$ to choose actions. The goal of each agent is to maximize its own expected return $G_i = \Sigma_{t=0}^{T}\gamma^tr_i(s_t, a_t^i)$ where $\gamma$ is a discount factor and $T$ denotes the time horizon.


\subsection{Delay-Aware Markov Decision Process}
The delay-free MDP is problematic with agent delays and could lead to arbitrarily suboptimal policies \cite{singh1994learning}. To retrieve the Markov property, Delay-Aware MDP (DA-MDP) is proposed~\cite{chen2020delayaware} by augmenting the state space $S$ to $\pmb{\mathcal{X}}$:
\begin{definition} \label{def:DMDP}
A Delay-Aware Markov Decision Process $D\!A\!M\!D\!P(E, k)=(\pmb{\mathcal{X}}, \pmb{\mathcal{A}}, \augm \rho, \augm p, \augm r)$ augments a Markov Decision Process $M\!D\!P(E) = (\mathcal{S}, \mathcal{A}, \rho, p, r)$, such that \\
(1) augmented state space $\pmb{\mathcal{X}} = \mathcal{S} \times \mathcal{A}^{k}$ where $k$ denotes the delay step, \\
(2) action space $\pmb{\mathcal{A}} = \mathcal{A}$, \\
(3) initial state distribution
\begin{equation*}
\begin{aligned}
\augm \rho(\x_0) = 
\augm \rho({s_0, a_0, \dots, a_{k-1}}) = \rho(s_0) \ \prod_{i=0}^{k-1}\delta(a_i - c_i),
\end{aligned}
\footnote{$\delta$ is the Dirac delta function. If $y \sim \delta(\cdot - x)$ then $y=x$ with probability one.}
\end{equation*}
where $(c_i)_{i=0:k-1}$ denotes the initial action sequence, \\
(4) transition distribution
\begin{equation*}
\begin{aligned}
&\augm p(\x_{t+1}|\x_t, \augm a_t) \\
&= 
\augm p({s_{t+1}, a_{t+1}^{(t+1)}, \dots, a_{t+k}^{(t+1)}} | {s_t, a_t^{(t)}, \dots, a_{t+k-1}^{(t)} }, \augm a_t) \\
&= p(s_{t+1} | s_t, a_t^{(t)}) \prod_{i=1}^{k-1}\delta(a_{t+i}^{(t+1)} - a_{t+i}^{(t)}) \delta(a_{t+k}^{(t+1)} - \augm a_t),
\end{aligned}
\end{equation*}
(5) reward function 
\begin{equation*}
\begin{aligned}
\augm r(\x_t, \augm a_t) = 
\augm r({s_t, a_t, \dots, a_{t+k-1}}, \augm a_t) = r(s_t, a_t).
\end{aligned}
\end{equation*}
The state vector $\pmb{x}$ of DA-MDP is augmented with an action sequence 
being executed in the next $k$ steps where $k \in \mathbb{N}$ is the delay duration.
The superscript of $a_{t_1}^{(t_2)}$ means that the action is an element of $\x_{t_2}$ and the subscript represents the action's executed time.
$\augm a_t$ is the action taken at time $t$ in a DA-MDP but executed at time $t+k$ due to the $k$-step action delay, i.e., $\augm a_t= a_{t+k}$.
\end{definition}

Policies interacting with the DA-MDPs, which also need to be augmented since the dimension of state vectors has changed, are denoted by $\augm \pi$. 


It should be noted that both action delay and observation delay could exist in real-world systems. However, it has been fully discussed and proved that from the perspective of the learning agent, observation and action delays form the same mathematical problem, since they both lead to the mismatch between the current observation and the executed action~\cite{katsikopoulos2003markov}.
For simplicity, we will focus on the action delay in this paper, and the algorithm and conclusions should be able to generalize to systems with observation delays. 

The above definition of DA-MDP assumes that the delay time of the agent is an integer multiple of the time step of the system, which is usually not true for many real-world tasks like robotic control. For that, Schuitema \textit{et al.} \cite{schuitema2010control} has proposed an approximation approach by assuming a \textit{virtual} effective action at each discrete system time step, which could achieve first-order equivalence in linearizable systems with arbitrary delay time. With this approximation, the above DA-MDP structure can be adapted to systems with arbitrary-value delays.

\subsection{Multi-Agent Deep Deterministic Policy Gradient}

Reinforcement learning has been used to solve Markov games. 
The simplest way is to directly train each agent with single-agent reinforcement learning algorithms. However, this approach will introduce the non-stationarity issue since the learning agent is not aware of the evolution of other agents that treated as part of the environment, thus violate the Markov property that is required for the convergence of most reinforcement learning algorithms \cite{tan1993multi, sutton2018reinforcement}. To alleviate the non-stationarity issue introduced by the multi-agent setting, several approaches have been proposed \cite{papoudakis2019dealing}. \textit{Centralized training and decentralized execution} is one of the most widely used diagram for multi-agent reinforcement learning. 
Lowe \textit{et al.} \cite{lowe2017multi} utilized this diagram and proposed the \textit{multi-agent deep deterministic policy gradient} (MADDPG) algorithm.
The core idea of MADDPG is to learn a centralized action-value function (\textit{critic}) and a decentralized policy (\textit{actor}) for each agent. The centralized \textit{critic} conditions on global information to alleviate the non-stationarity problem, while the decentralized \textit{actor} conditions only on private observation to avoid the need for a centralized controller during execution.



A brief description of MADDPG is as follows. In a game with $N$ agents, let $\augm \cpol = \{\cpol_1, ..., \cpol_N\}$ be the set of all agent policies parameterized by $\pmb{\theta} = \{\theta_1, ..., \theta_N\}$, respectively. Based on the the deterministic policy gradient (DPG) algorithm \cite{lillicrap2015continuous}, we can write the gradient of the objective function $J(\theta_i) = \mathbb{E}[G_i]$ for agent $i$ as:
\begin{equation}
\label{equ:dpg}
\begin{aligned}
&\nabla_{\theta_i} J(\theta_i)=  \\
&\mathbb{E}_{x,\augm a \sim \mathcal{D}}[\nabla_{\theta_i} \cpol_i(a_i|o_i) \nabla_{a_i} Q^{\augm \cpol}_i (x,a_1, ..., a_N)|_{a_i=\cpol_i (o_i)}].
\end{aligned}
\end{equation}


In Equ.~\ref{equ:dpg}, $Q^{\augm \cpol}_i (x,a_1, ..., a_N)$ is the centralized Q function (\textit{critic}) for agent $i$ that conditions on the global information including the global state representation ($x$) and the actions of all agents ($a_1,\ldots, a_N$). Under this setting, agents can have different reward functions since each $Q^{\augm \cpol}_i$ is learned separately, which means this algorithm can be used in both cooperative and competitive tasks.

Based on deep Q-learning \cite{mnih2013playing}, the centralized Q function $Q^{\cpol}_i$ for agent $i$ is updated as:
\begin{align*}
\mathcal{L}(\theta_i) &= \mathbb{E}_{x,\augm a,\augm r,x'}[(Q^{\augm \cpol}_i(x,a_1,\ldots,a_N) - y)^2], \\
\text{where     } y &= r_i + \gamma\, Q^{\augm \cpol'}_i(x', a_1',\ldots,a_N')\big|_{a_j'=\cpol'_j(o_j)}.
\end{align*}
Here, $\augm \cpol' = \{\cpol_{\theta'_1}, ..., \cpol_{\theta'_N} \}$ is the set of target policies with soft-updated parameters $\theta'_i$ to stabilize training \cite{mnih2013playing}.

\section{Delay-Aware Markov Game}
\label{sec:damg}

Ignoring delays violates the Markov property in multi-agent scenarios and could lead to arbitrarily suboptimal policies. To retrieve the Markov property, we formally define the Delay-Aware Markov Game (DA-MG) as below:
\begin{definition} \label{def:DAMG}
A Delay-Aware Markov Game with $N$ agents
$D\!A\!M\!G(E, \augm k)=(\pmb{\mathcal{X}}, \pmb{\mathcal{A}}, \pmb{\mathcal{O}}, \augm \rho, \pmb{p}, \augm r)$ augments a Markov Game $M\!G(E) = (\mathcal{S}, \mathcal{A}, \mathcal{O}, \rho, p,  r)$, such that \\
(1) augmented state space $\pmb{\mathcal{X}} = \mathcal{S} \times \mathcal{A}_{1}^{k_1} \times \dots \times \mathcal{A}_{N}^{k_N}$ where $k_i$ denotes the delay step of agent $i$,\\
(2) action space $\pmb{\mathcal{A}} = \mathcal{A}$, \\
(3) initial state distribution
\begin{equation*}
\begin{aligned}
\augm \rho(\x_0) &= \augm \rho({s_0, a^1_0, \dots, a^1_{k_1-1}, \dots, a^N_0, \dots, a^N_{k_N-1}}) \\
&= \rho(s_0) \ \prod_{i=1}^{N}\prod_{j=0}^{k_i-1}\delta(a^i_j - c^i_j),
\end{aligned}
\end{equation*}
where $(c^i_j)_{i=1:N, j=0:k_i-1}$ denotes the initial action sequences of all agents,\\
(4) transition distribution
\begin{equation*}
\begin{aligned}
&\pmb{p} (\x_{t+1}|\x_t, \augm a_t) \\
= &\pmb{p}({s_{t+1}, a^{1, (t+1)}_{t+1}, \dots, a^{1, (t+1)}_{t+k_1},\dots, a^{N, (t+1)}_{t+1}, \dots, a^{N, (t+1)}_{t+k_N} } \\
&| s_t, a^{1, (t)}_{t}, \dots, a^{1, (t)}_{t+k_1-1},\dots, a^{N, (t)}_{t}, \dots, a^{N, (t)}_{t+k_N-1}, \augm a_t) \\
= & p(s_{t+1} | s_t, a^{1, (t)}_t, \dots, a^{N, (t)}_t) \\
&\prod_{i=1}^{N}\prod_{j=1}^{k_i-1}\delta(a^{i, (t+1)}_{t+j} - a^{i, (t)}_{t+j}) \prod_{i=1}^{N} \delta(a^{i, (t+1)}_{t+k_i} - \augm a_t^i)
\end{aligned}
\end{equation*}
(5) reward function 
\begin{equation*}
\begin{aligned}
\augm r_i(\x_t, \augm a_t) = r_i(s_t, a^i_t)
\end{aligned}
\end{equation*}
for each agent $i$.
\end{definition}

DA-MGs have an augmented state space $\mathcal{S} \times \mathcal{A}_{1}^{k_1} \times \dots \times \mathcal{A}_{N}^{k_N}$.  $k_i$ denotes the delay step of agent $i$.
$a_{t_1}^{i,(t_2)}$ is an element of  $\x_{t_2}$ and denotes the action of agent $i$ executed at time $t_1$.
$\augm a_t$ is the action vector taken at time $t$ in a DA-MG; its $i$-th element $\augm a_t^i$  is  executed by agent $i$ at time $t+k_i$ due to the $n_i$-step action delay, i.e. $\augm a_t^i= a^i_{t+k_i}$.
Policies interacting with the DA-MDPs, which also need to be augmented since the dimension of state vectors has changed, are denoted by $\augm \pi$. 

\begin{figure}[h]
  \centering
  \includegraphics[width=0.9\linewidth]{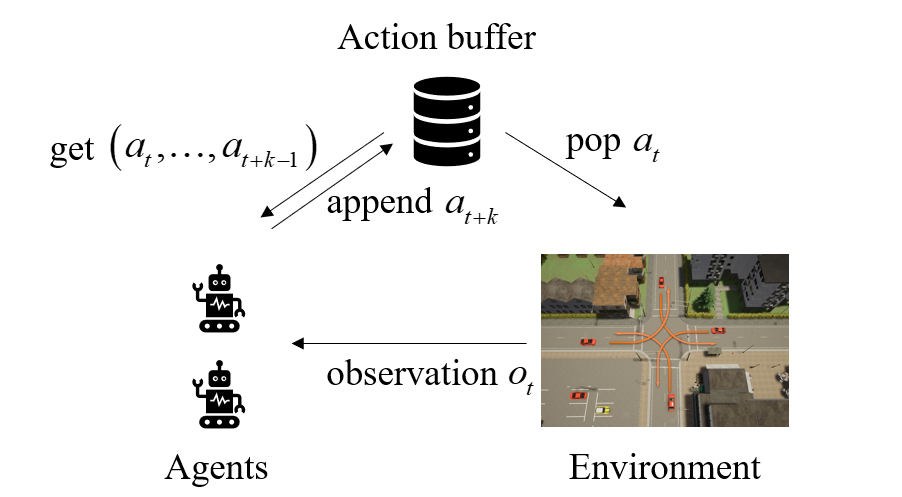}
  \caption{Interaction manner between delayed agents and the environment. The agent interact with the environment not directly but through an action buffer. At time $t$, agents get the observation $o_t$ from the environment as well as a future action sequences $(a_t, \dots, a_{t+k-1})$ from the action buffer. The agents then decide their future action $a_{t+k}$ and store them in the action buffer. The action buffer then pops actions $a_t$ to be executed to the environment.}
  \label{fig:actbuffer}
\end{figure}

To prove the solidity of Definition~\ref{def:DAMG}, we need to show that a Markov game with multi-step action delays can be converted to a regular Markov game by state augmentation (DA-MG). We prove the equivalence of these two by comparing their corresponding Markov Reward Processes (MRPs).
The delay-free MRP for a Markov Games is:

\begin{definition} \label{def:MRP}
A Markov Reward Process $(\mathcal{S}, \rho, \kappa, \bar r) = M\!R\!P(M\!G(E), \augm \pi)$ can be derived from a Markov Game $M\!G(E) =(\mathcal{S}, \mathcal{A}, \mathcal{O}, \rho, p,  r)$ with a set of policy $\augm \pi = \{\pi_1, \dots, \pi_N\}$, such that
\begin{equation*} \label{mdp_kernel}
\kappa(s_{t+1} | s_t) = \int_\mathcal{A} p(s_{t+1} | s_t, a^1_t, \dots, a^N_t) \prod_{i=1}^{N}\left[ \pi_i(a^i_t | o_t^i) \ d a_t^i\right],
\end{equation*}
\begin{equation*} \label{mdp_reward}
\bar r_i(s_t) = \int_{\mathcal{A}_i} r_i(s_t, a^i_t) \pi_i(a^i_t | o_t^i) \ d a_t^i ,
\end{equation*}
for each agent $i$. $\kappa$ is the state transition distribution and $\bar r$ is the state reward function of the MRP. $E$ is the original environment without delays.
\end{definition}
In the delay-free framework, at each time step, the agents select actions based on their current observations. The actions will immediately be executed in the environment to generate the next observations.
However, if the action delay exists, the interaction manner between the environment and the agents changes, and a different MRP will be generated. An illustration of the delayed interaction between agents and the environment is shown in Fig.~\ref{fig:actbuffer}. The agents interact with the environment not directly but through an action buffer.

Based on the delayed interaction manner between the agents and the environment, the Delay-Aware MRP (DA-MRP) is defined as below.

\begin{definition} \label{def:DAMRP}
A Delay-Aware Markov Reward Process with $N$ agents $(\pmb{\mathcal{X}}, \augm \rho, \pmb{\kappa}, \pmb{\bar r}) = D\!A\!M\!R\!P(M\!G(E), \augm \pi, \augm k)$ can be derived from a Markov Game $M\!G(E) = (\mathcal{S}, \mathcal{A}, \mathcal{O}, \rho, p,  r)$ with a set of policy $\augm \pi = \{\augm \pi_1, \dots, \augm \pi_N\}$ and a set of delay step $\augm k = \{k_1, \dots, k_N\}$, such that \\
(1) augmented state space
\begin{equation*}
    \pmb{\mathcal{X}} = \mathcal{S} \times \mathcal{A}_{1}^{k_1} \times \dots \times \mathcal{A}_{N}^{k_N},
\end{equation*}
(2) initial state distribution
\begin{equation*}
\begin{aligned}
\augm \rho(\x_0) =& \augm \rho({s_0, a^1_0, \dots, a^1_{k_1-1}, \dots, a^N_0, \dots, a^N_{k_N-1}}) \\
=& \rho(s_0) \ \prod_{i=1}^{N}\prod_{j=0}^{k_i-1}\delta(a^i_j - c^i_j),
\end{aligned}
\end{equation*}
where $(c^i_j)_{i=1:N, j=0:k_i-1}$ denotes the initial action sequences of all agents, \\
(3) state transition distribution 
\begin{equation*}
\begin{aligned}
& \augm \kappa\left(\x_{t+1}|\x_t\right)  \\
=& \augm \kappa({s_{t+1}, a^{1, (t+1)}_{t+1}, \dots, a^{1, (t+1)}_{t+k_1},\dots, a^{N, (t+1)}_{t+1}, \dots, a^{N, (t+1)}_{t+k_N} } \\
&| s_t, a^{1, (t)}_{t}, \dots, a^{1, (t)}_{t+k_1-1},\dots, a^{N, (t)}_{t}, \dots, a^{N, (t)}_{t+k_N-1}) \\
=&p(s_{t+1} | s_t,  a^{1, (t)}_t, \dots, a^{N, (t)}_t) \\
&\prod_{i=1}^{N}\prod_{j=1}^{n_i-1}\delta(a^{i, (t+1)}_{t+j} - a^{i, (t)}_{t+j})  \prod_{i=1}^{N} \augm \pi_i(a^{i, (t+1)}_{t+k_i} | \augm o_{t}^i) 
,
\end{aligned}
\end{equation*}
(4) state-reward function 
\begin{equation*}
\begin{aligned}
\pmb{\bar r_i}({\x_t})= r_i(s_t, a^{i}_{t})
\end{aligned}
\end{equation*}
for each agent $i$.

The input of policy for agent $i$ at time $t$ has two parts: $\augm o_t^i = (o_{t, obs}^i, o_{t, act}^i)$, where $o_{t, obs}^i$ is the observation of the environment and $o_{t, act}^i$ is a planned action sequence for agent $i$ of length $k_i$ that will be executed from current time step: $o_{t, act}^i = (a_t^i, \dots, a_{t+k_i-1}^i)$.

\end{definition}

With Def.~\ref{def:MG}-~\ref{def:DAMRP}, we are ready to prove that DA-MG is a correct augmentation of MG with delay, as stated in Theorem~\ref{the:equn}.
\begin{restatable}{theorem}{TheoremRttb2}
\label{the:equn}
A set of policy $\augm \pi: \mathcal{A} \times \pmb{\mathcal{X}} \to [0,1]$ interacting with $D\!A\!M\!G(E, \augm k)$ in the delay-free manner produces the same Markov Reward Process as $\augm \pi$ interacting with $M\!G(E)$ with $\augm k$ action delays for agents, i.e.
\begin{equation} \label{DMDP_equality}
    D\!A\!M\!R\!P(M\!G(E), \augm \pi, \augm n) = M\!R\!P(D\!A\!M\!G(E, \augm n), \augm \pi).
\end{equation}
\end{restatable}
The proof is provided in the Appendix.

\begin{figure*}[t]
  \centering
  \includegraphics[width=0.7\linewidth]{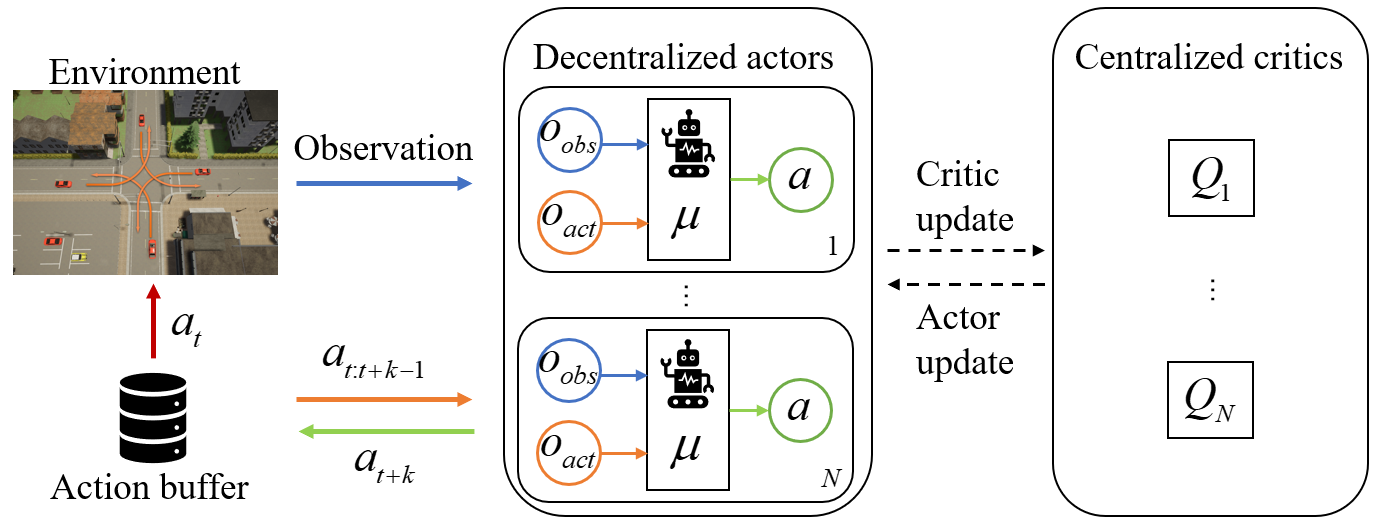}
  \caption{The framework of \textit{Delay-Aware Multi-Agent Reinforcement Learning} (DAMARL). We adopt the paradigm of centralized training with decentralized execution: for each agent $i$, we learn a centralized action-value function $Q_i$ (\textit{critic}) which conditions on global information and a decentralized policy $\mu_i$ (\textit{actor}) that only needs partial observation. For each agent, the input of agent policy has two parts: $o = (o_{obs}, o_{act})$, where $o_{obs}$ is the observation of the environment and $o_{act}$ is a planned action sequence that will be executed from current time step.  }
  \label{fig:damaddpg}
\end{figure*}

\section{Delay-Aware Multi-agent Reinforcement Learning}
\label{sec:damaddpg}
Theorem.~\ref{the:equn} shows that instead of solving MGs with delays, we can alternatively solve the corresponding DA-MGs directly with DRL. Based on this finding, we proposed the framework of \textit{Delay-Aware Multi-Agent Reinforcement Learning} (DAMARL).
To alleviate the non-stationarity issue introduced by the multi-agent setting, we adopt the paradigm of centralized training with decentralized execution: for each agent,
we learn a centralized Q function (\textit{critic})
which conditions on global information
and a decentralized policy (\textit{actor}) that only needs partial observation. 
An illustration of the framework is shown in Fig.~\ref{fig:damaddpg}.
The main advantages of this structure are as follows:
\begin{itemize}
    \item The non-stationarity problem is alleviated by centralized training since the transition distribution of the environment is stationary when knowing all agent actions.
    \item A centralized controller to direct all agents, which is not realistic to deploy in many real-world multi-agent scenarios, is not needed with decentralized policies.
    \item We learn an individual Q function for each agent, allowing them to have different reward functions so that we can adopt this algorithm in both cooperative and competitive multi-agent tasks.
    \item Individual Q functions and policies allow agents to have different delay steps.
\end{itemize}

With the framework of DAMARL, we can adapt any DRL algorithm with the actor-critic structure \cite{sutton2018reinforcement} to a delay-aware algorithm such as Advantage Actor-Critic (A2C) \cite{mnih2016asynchronous}, Deep Deterministic Policy Gradient (DDPG) \cite{lillicrap2015continuous} and Soft Actor-Critic (SAC) \cite{haarnoja2018soft}. In this paper, we update the multi-agent version of DDPG with delay-awareness and propose \textit{delay-aware multi-agent deep deterministic policy gradient} (DAMA-DDPG).
 Concretely, in a game with $N$ agents, let $\augm \cpol = \{\cpol_1, ..., \cpol_N\}$ be the set of all agent policies parameterized by $\pmb{\theta} = \{\theta_1, ..., \theta_N\}$, respectively. Based on the the deterministic policy gradient (DPG) algorithm \cite{lillicrap2015continuous}, we can write the gradient of the objective function $J(\theta_i) = \mathbb{E}[G_i]$ for agent $i$ as:
\begin{equation}
\label{equ:dadpg}
\begin{aligned}
&\nabla_{\theta_i} J(\augm \cpol_i) =\\
&\mathbb{E}_{\x,\augm a \sim \mathcal{D}}[\nabla_{\theta_i} \augm \cpol_i(\augm a_i|\augm o_i) \nabla_{\augm a_i} Q^{\augm \cpol}_i (\augm x,\augm a_1, ..., \augm a_N)|_{\augm a_i=\augm \cpol_i (\augm o_i)}],
\end{aligned}
\end{equation}
The structure of Eq.~\ref{equ:dadpg} is in conformity with the original deterministic policy gradient (Eq.~\ref{equ:dpg}). However, the policies $\augm \mu$,  states $\augm x$ and observations $\augm o$ are augmented based on the DA-MG proposed in Def.~\ref{def:DAMG}. 
In Equ.~\ref{equ:dadpg}, $Q^{\augm \cpol}_i (\augm x,\augm a_1, ..., \augm a_N)$ is the centralized Q function (\textit{critic}) for agent $i$ that conditions on the global information including the global state representation ($\x$) and the actions of all agents ($\augm a_1, ..., \augm a_N$).
In the delay-aware case, $\x$ could consist of the observations of all agents as well as action sequences of all agents in a near future, $\augm x = (\augm o_1, ...,\augm o_N)$, where $\augm o_i$ is the input of the policy $\augm \mu_i$ of agent $i$ and has two parts: $\augm o_i = (o_{obs}^i,  o_{act}^i)$. Here, $o_{obs}^i$ is the observation of the environment by the $i$-th agent, and  $o_{act}^i$ is a planned action sequence for agent $i$ of length $k_i$ that will be executed from current time step.
For example, at time $t$, $o_{t, act}^i = a_{t:t+k_i-1}^i$. The $o_{act}^i$ is fetched from an action buffer that serves as a bridge between  the agents and the environment, as shown in Fig.~\ref{fig:actbuffer}.

The replay buffer $\mathcal{D}$ is used to record historical experiences of all agents. 
The centralized Q function $Q^{\augm \cpol}_i$ for agent $i$ is updated as:
\begin{equation*}
\begin{aligned}
\mathcal{L}(\theta_i) = \mathbb{E}_{\x,\augm a,\augm r,\x'}[(Q^{\augm \cpol}_i(\x,\augm a_1,\dots,\augm a_N) - y)^2], \\
\text{where   } y = r_i + \gamma\, Q^{\augm \cpol'}_i(\x', \augm a_1',\dots,\augm a_N')\big|_{\augm a_j'=\augm \cpol'_j(\augm o_j)}.
\end{aligned}
\end{equation*}
Here, $\augm \cpol' = \{\augm \cpol_{\theta'_1}, \dots, \augm \cpol_{\theta'_N} \}$ is the set of augmented target policies with soft-updated parameters $\theta'_i$ used to stabilize training.



The description of the full algorithm is shown in Algorithm~\ref{alg:damaddpg}.

\begin{algorithm}[t]
  \begin{algorithmic}
  \STATE Initialize the experience replay buffer $\mathcal{D}$
    \FOR{$\textrm{episode}=1\textrm{ to }M$}
      \STATE Initialize the action noise $\mathcal{N}_t$ and the action buffer $\mathcal{F}$
      \STATE Get initial state $\x_0$
      \FOR{$t=1\textrm{ to }T$}
            \FOR{agent $i=1\textrm{ to }N$}
                \STATE get $\augm o_i = (o^i_{obs}, o^i_{act})$ from the environment and $\mathcal{F}$
               \STATE select action $\augm a_i=\augm \cpol_{\theta_i}(\augm o_i)+\mathcal{N}_t$
               \ENDFOR
        \STATE Store actions $\augm a=(\augm a_1,\dots,\augm a_N)$ in $\mathcal{F}$
        \STATE Pop $\augm a=( a_1,\dots, a_N)$ from $\mathcal{F}$ and execute it
        \STATE get the reward $\augm r$ and the new state $\augm x'$
          \STATE Store $(\x,\augm a,\augm r,\x') \rightarrow \mathcal{D}$
          \STATE $\x \gets\x'$
          \FOR{agent $i=1\textrm{ to }N$}
            \STATE Randomly sample a batch of $B$ samples $(\x^b,\augm a^b,\augm r^b, \x'^b)$ from $\mathcal{D}$
            \STATE Set $y^b=r_i^b+\gamma\, Q_i^{\augm \cpol'}(\x'^b,\augm a_1',\dots,\augm a_N')|_{\augm a_l'=\augm \cpol'_l(\augm o_l^b)}$
            \STATE Update centralized critics with loss $\mathcal{L}(\theta_i)=\frac{1}{B}\sum_j\left(y^b-Q_i^{\augm \cpol}(\x^b,\augm a^b_1,\dots, \augm a^b_N)\right)^2$
            \STATE Update decentralized actors by
            $
            \nabla_{\theta_i}J \approx \frac{1}{B}\sum_j\nabla_{\theta_i}\augm \cpol_i(\augm o_i^b)\nabla_{a_i}Q_i^{\augm \cpol}(\x^b,\augm a^b_1,\dots,\augm a^b_N)\big|_{\augm a_i=\augm \cpol_i(\augm o_i^b)}
            $
          \ENDFOR     
          \STATE Soft update of target networks for each agent $i$:
          $$
          \theta_i'\gets\kappa\theta_i+(1-\kappa)\theta_i'
          $$
      \ENDFOR
    \ENDFOR
  \end{algorithmic}
 \caption{DAMA-DDPG}
\label{alg:damaddpg}
\end{algorithm}

\section{Experiment}
\label{sec:experiment}
To show the performance of DAMARL, we adopt two environment platforms. One is the multi-agent particle environment platform proposed in \cite{lowe2017multi}\footnote{https://github.com/openai/multiagent-particle-envs} where the agents are particles that move on a two-dimensional plane to achieve cooperative or competitive tasks. The other is revised based on a traffic platform Highway~\cite{highway-env} where we construct an unsignalized intersection scenario that requires coordination of all road users. Implementation details and demo videos are provided in our GitHub respiratory\footnote{https://github.com/baimingc/delay-aware-MARL}. Important hyper-parameters are shown in Table~\ref{tab:hyper}.

\begin{table}[h]
\centering
\caption{Hyper-parameters}\smallskip
\label{tab:hyper}
\begin{tabular}{c c} 
    \hline
    Description & Value \\
    \hline
    learning rate & 0.01 \\
    discount factor $\gamma$ & 0.99 \\
    soft update coefficient $\kappa$ & 0.01 \\
    replay buffer size & $10^6$ \\
    batch size $B$ & 1024\\
    episode length $T$ & 25\\
    \hline
\end{tabular}
\end{table}

\begin{figure*}[t]
  \centering
  \begin{subfigure}[t]{0.3\linewidth}
    \centering\includegraphics[width=\linewidth]{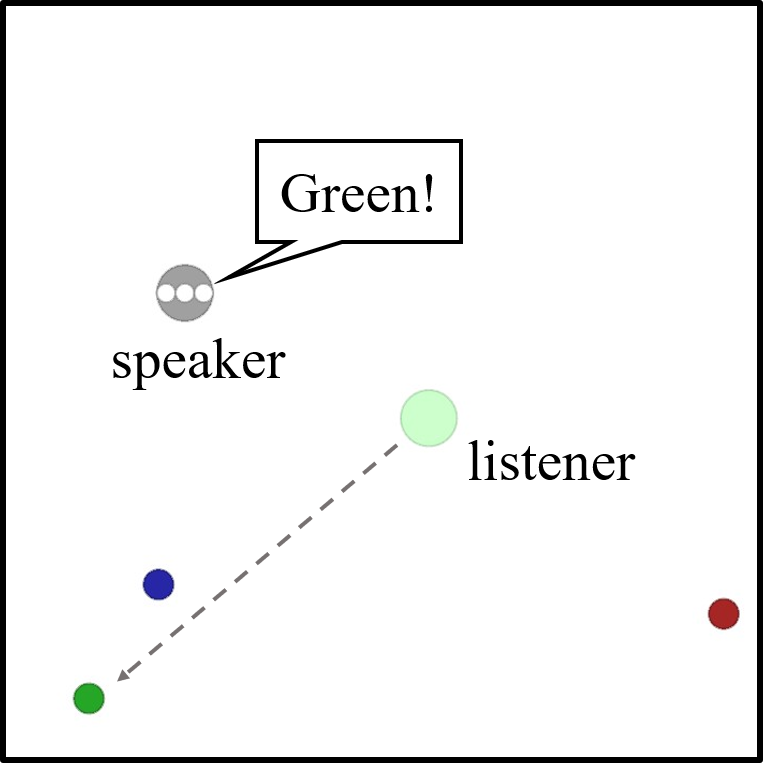}
    \caption{Cooperative communication}
  \end{subfigure}%
  \hfill
  \begin{subfigure}[t]{0.3\linewidth}
    \centering\includegraphics[width=\linewidth]{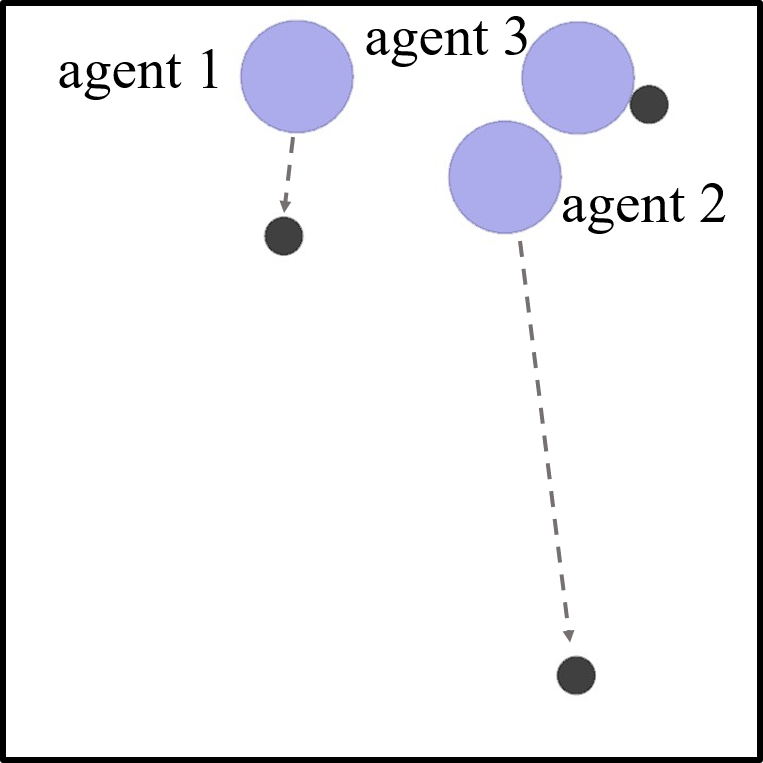}
    \caption{Cooperative navigation}
  \end{subfigure}
  \hfill
  \begin{subfigure}[t]{0.3\linewidth}
    \centering\includegraphics[width=\linewidth]{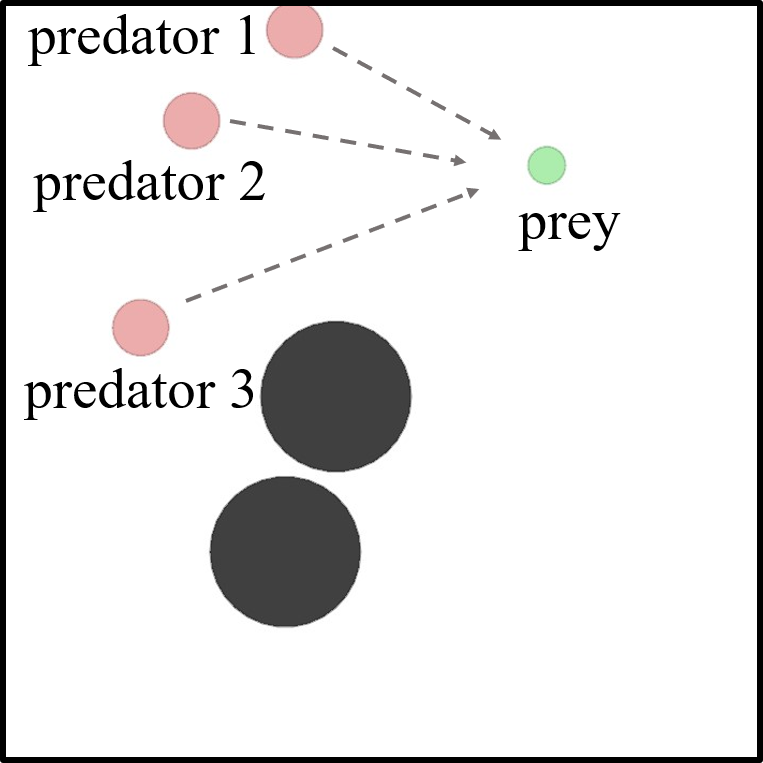}
    \caption{Predator-prey}
  \end{subfigure}
  \caption{Tasks in the multi-agent particle environment.}
  \label{fig:multiparticle}
\end{figure*}

\subsection{Multi-Agent Particle Environment}
The multi-agent particle environment is composed of several agents and landmarks in a two-dimensional world with continuous state space. Agents can move in the environment and send out messages that can be broadcasted to other agents. Some tasks are cooperative where all agents share one mutual reward function, while others are competitive or mixed where agents have inverse or different reward functions. In some tasks, agents need to communicate to achieve the goal, while in other tasks agents can only perform movements in the two-dimensional plane. We provide details for the used environments below. An illustration of the tasks is shown in Fig.~\ref{fig:multiparticle}.

\textbf{Cooperative communication.} Two cooperating agents are involved in this task, a speaker and a listener. They are spawned in an environment with three landmarks of different colors. The goal of the listener is to navigate to a landmark of a particular color. However, the listener can only observe the relative position and color of the landmarks, excluding which landmark it must navigate to. On the contrary, the speaker knows the color of the goal landmark, and it can send out a message at each time step which is heard by the listener. Therefore, to finish the cooperative task, the speaker and the listener must learn to communicate so that the listener can understand the message from the speaker and navigate to the landmark with the correct color.


\textbf{Cooperative navigation.} In this environment, three agents must collaborate to 'cover' all of the three landmarks in the environment by movement. In addition, these agents occupy a large physical space and are punished when they collide with each other. Agents can observe the position and speed of all agents as well as the position of all land marks.  



\textbf{Predator-prey.} In this environment, three slower cooperative predators must catch up with a faster prey in a randomly generated environment, with two large landmarks blocking the way. Each time a collaborating predator collides with the prey, the predators will be rewarded and the prey will be punished. The agents can observe the relative position and speed of other agents as well as the positions of the landmarks.


\begin{figure*}[t]
  \centering
  \begin{subfigure}[t]{0.8\linewidth}
   \centering\includegraphics[width=\linewidth]{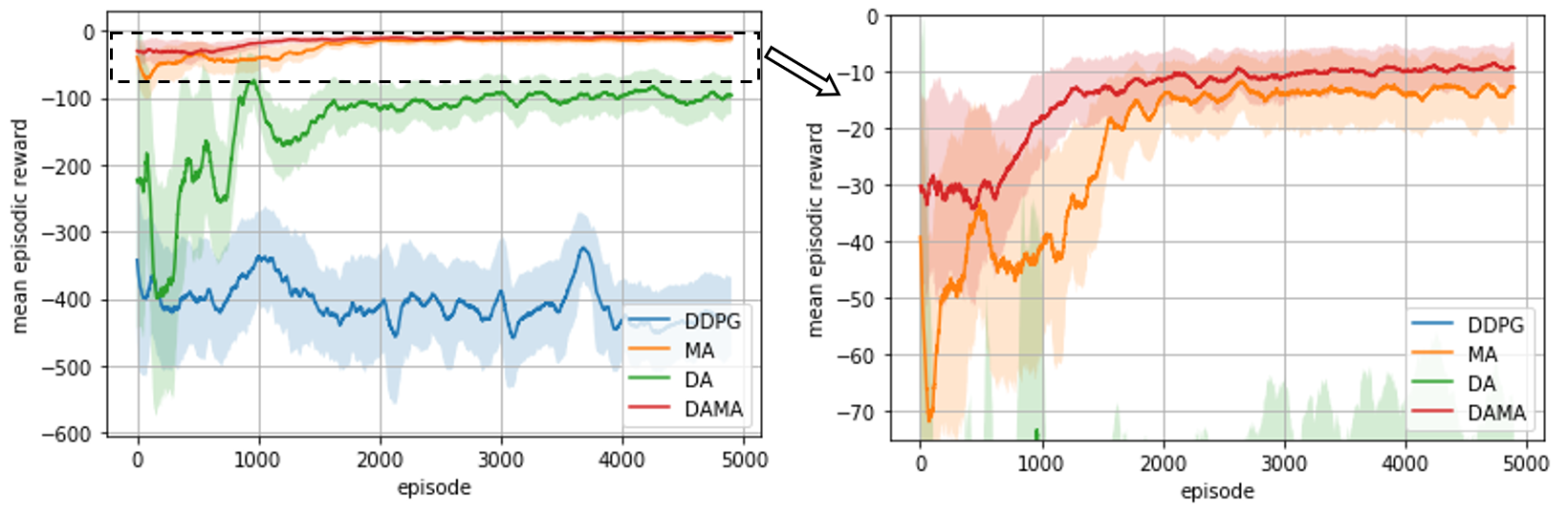}
    \caption{Cooperative communication ($\Delta t$ = 0.1 s)}
    \label{fig:coop:a}
  \end{subfigure}%
\vskip\baselineskip
  \begin{subfigure}[t]{0.4\linewidth}
    \includegraphics[width=\linewidth]{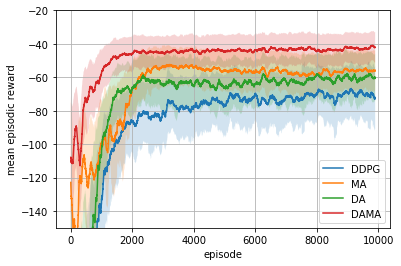}
    \caption{Cooperative navigation ($\Delta t$ = 0.2 s)}
    \label{fig:coop:c}
  \end{subfigure}
    \begin{subfigure}[t]{0.4\linewidth}
    \includegraphics[width=\linewidth]{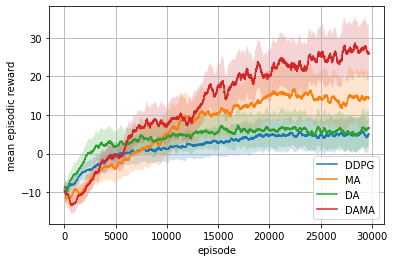}
    \caption{Predator-prey with fixed prey policy ($\Delta t$ = 0.2 s)}
    \label{fig:coop:d}
  \end{subfigure}
  \caption{Effect of delay-awareness. DAMA is the proposed algorithm that utilizes delay-awareness as well as multi-agent centralized training. It is clearly shown that DAMA outperforms other algorithms in all 3 tasks while the vanilla DDPG has the worst performance.}
  \label{fig:coop}
\end{figure*}

\begin{table*}[t]
\centering
\small
\caption{Number of touches in predator-prey ($\Delta t = 0.2$ s)}
\label{tab:pp}
\begin{tabular}{| c | c | c c c c | c c c| } 
\hline
\multirow{2}{*}{Delay time (s)}&\multirow{2}{*}{Algorithm of prey} &\multicolumn{4}{c|}{Algorithm of predators} &  \multicolumn{3}{c|}{Improvement of predators}\\
& & DAMA & MA & DA & DDPG & DAMA & MA & DA\\ 
\hline
\multirow{4}{*}{0.2} &DAMA& \textbf{10.3 $\pm$ 2.1} & 9.6 $\pm$ 2.0 & 8.5 $\pm$ 1.9 & 8.1 $\pm$ 1.8 & \textbf{2.2} & \textbf{1.5}& 0.4 \\ 
&MA& \textbf{12.1 $\pm$ 2.3} & 10.1 $\pm$ 2.1 & 9.0 $\pm$ 2.1 & 8.5 $\pm$ 1.8 & \textbf{3.6} & \textbf{1.6} & 0.5 \\ 
&DA& \textbf{15.8 $\pm$ 2.9} & 13.2 $\pm$ 2.7 & 9.7 $\pm$ 2.0& 8.8 $\pm$ 1.8 & \textbf{7.0} & \textbf{4.4} & 0.9\\ 
&DDPG& \textbf{17.0 $\pm$ 3.2} & 14.9 $\pm$ 2.9 & 11.4 $\pm$ 2.2 & 9.1 $\pm$ 1.9 & \textbf{7.9} & \textbf{5.8} & 2.3\\ 
\hline
\multirow{4}{*}{0.4}&DAMA& \textbf{10.1 $\pm$ 1.9} & 8.6 $\pm$ 1.7 & 8.2 $\pm$ 1.7 & 7.3 $\pm$ 1.6 & \textbf{2.8} & \textbf{1.3} & 0.9\\ 
&MA& \textbf{14.2 $\pm$ 2.8} & 9.5 $\pm$ 2.0 & 9.0 $\pm$ 1.8 & 7.9 $\pm$ 1.8 & \textbf{6.3} & \textbf{1.6} & 1.1\\ 
&DA& \textbf{14.9 $\pm$ 2.9} &10.7 $\pm$ 2.2 & 9.5 $\pm$ 1.9& 8.2 $\pm$ 1.8& \textbf{6.7} &\textbf{2.5} & 1.3\\ 
&DDPG& \textbf{17.6 $\pm$ 3.3} & 14.5 $\pm$ 2.9 & 13.1 $\pm$ 2.5 & 8.8 $\pm$ 1.9 & \textbf{8.8} & \textbf{5.7} & 4.3\\ 
\hline
\multirow{4}{*}{0.6}&DAMA& \textbf{9.6 $\pm$ 1.9} & 6.9 $\pm$ 1.7 & 7.6 $\pm$ 1.7 & 5.9 $\pm$ 1.6 & \textbf{3.7} & 1.0 & \textbf{1.7}\\ 
&MA& \textbf{16.0 $\pm$ 3.1} & 8.8 $\pm$ 2.0 & 12.5 $\pm$ 2.4 & 7.4 $\pm$ 1.7 & \textbf{7.6} & 1.4 &\textbf{5.1}\\ 
&DA& \textbf{13.7 $\pm$ 2.9} & 7.5 $\pm$ 1.7 & 9.3 $\pm$ 1.9& 6.2 $\pm$ 1.6& \textbf{7.5} & 1.3 & \textbf{3.1}\\ 
&DDPG& \textbf{19.2 $\pm$ 3.8} & 13.5 $\pm$ 2.8 & 16.8 $\pm$ 3.4 & 8.3 $\pm$ 1.8 & \textbf{10.9} & 5.2 & \textbf{8.5}\\ 
\hline
\end{tabular}
\end{table*}


\begin{figure*}[t]
  \centering
  \includegraphics[width=0.8\linewidth]{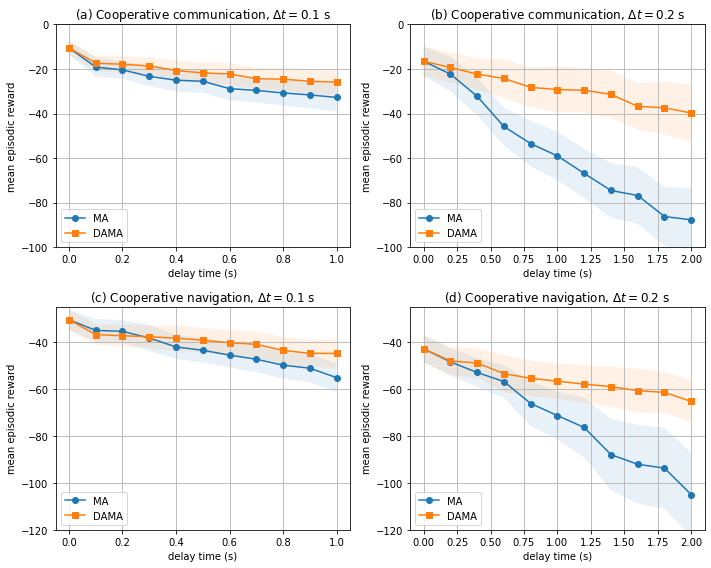}
  \caption{Performance of delay-aware (DAMA) and delay-unaware (MA) algorithms in cooperative communication and cooperative navigation scenarios with different agent delay times. As the delay time increases, both DAMA and MA algorithms get degraded performance. In most cases, the DAMA algorithm maintains higher performance than the MA algorithm. The performance gap gets more significant as the delay time increases.}
  \label{fig:sen}
\end{figure*}


\subsubsection{Effect of Delay-Awareness}
To show the effect of delay-awareness, we first test our algorithm on cooperative tasks including cooperative communication, cooperative navigation, and predator-prey where we adopt a fixed prey policy and only train the cooperative predators. To support discrete actions used for message communication in the cooperative communication task, we use the Gumbel-Softmax estimator \cite{jang2016categorical}. Unless specified, our policies and Q functions are parameterized by two-layer neural networks with 128 units per layer. Each experiment is run with 5 random seeds. The baseline algorithms are DDPG and MA-DDPG that use decentralized and centralized training, respectively, without delay-awareness.
We test the proposed delay-aware algorithm DAMA-DDPG (Algorithm~\ref{alg:damaddpg}). We also adapt DAMA-DDPG to Delay-Aware DDPG (DA-DDPG) which uses decentralized training and test it for comparison. 

For simplicity, we omit '-DDPG' when referring to an algorithm throughout the experiment part since our framework can be adapted to any DRL algorithms with the actor-critic structure. For example, 'DAMA-DDPG' is shortened by 'DAMA'.


The performance of the aforementioned algorithms in cooperative tasks indicated by episodic reward is shown in Fig.~\ref{fig:coop}. We use $\Delta t$ to denote the simulation timestep which is 0.1 seconds for cooperative communication and 0.2 for cooperative navigation and predator-prey. The agents are with 1-step action delay in all tasks ($k_i=1$ for $i = 1, \dots, N$).
We will change the simulation timestep as well as the agent delay time in the later part of the environment. It is clearly shown that DAMA outperforms other algorithms in all 3 tasks with delay-awareness and centralized training, while the vanilla DDPG has the worst performance. The result of cooperation communication (Fig.~\ref{fig:coop:a}) shows the importance of centralized training. In this task, the action of the speaker significantly affects the behavior of the listener by setting the goal, so a centralized Q function that conditions on all agent actions will greatly stabilize training. The advantage of delay-awareness is more significant in the high-dynamic task, predator-prey, where a prey is running fast to escape from the agents (Fig.~\ref{fig:coop:d}).

We also test our algorithm in a competitive task: predator-prey. To compare performance, we train agents and adversaries with different algorithms and let them compete with each other. The simulation timestep $\Delta t$ is set to 0.2 seconds. The delay times of agents are 0.2 s, 0.4 s and 0.6 s in each set of experiments. We evaluate the performance of the aforementioned algorithms by the number of prey touches by predators per episode. Since the goal of predators is to touch the prey as many times as possible, a higher number of touches indicates a stronger predator policy against a weaker prey policy.
The results on the predator-prey task are shown in Table~\ref{tab:pp}. All agents are trained with 30,000 episodes. It is clearly shown that DAMA has the best performance against other algorithms: 
with any delay time, the DAMA predator gets the highest touch number against the DDPG prey, while the DDPG predator gets the lowest touch number against the DAMA prey. Also, as the delay time increases, the delay-awareness gets more important than multi-agent centralized training, as shown in the last column of Table~\ref{tab:pp}. When the delay time is relatively small as 0.2 s, the improvement of predator policies by utilizing multi-agent centralized training is larger than delay-awareness. When the delay time grows to 0.6 s, however, the situation is reversed and delay-awareness has a larger impact on the improvement of predator policies.



\subsubsection{Delay Sensitivity}

Sensitivity of delay is another important metric to evaluate reinforcement learning algorithms. Delay-aware algorithms may experience performance degradation due to state-space expansion. To show the value of delay-awareness, we compare the performance of the aforementioned delay-aware (DAMA) and delay-unaware (MA) algorithms with different agent delay times. We perform experiments in cooperative communication and cooperative navigation scenarios. Results are shown in Fig.~\ref{fig:sen}. The agent delay step is $k_i = 0, 1, \dots, 10$ in each sub-figure.
The simulation timestep $\Delta t$ is 0.1 seconds in Fig.~\ref{fig:sen}a and~\ref{fig:sen}c and 0.2 seconds in Fig.~\ref{fig:sen}b and~\ref{fig:sen}d. It is shown that as the delay time increases, both delay-aware and delay-unaware algorithms get degraded performance. In most cases, the delay-aware algorithm maintains higher performance than the delay-unaware algorithm, and the performance gap gets more significant as the delay time increases. 

The only exception is in Fig.~\ref{fig:sen}c when the delay time is less than 0.2 seconds. The performance of the delay-unaware algorithm is slightly better than the delay-aware algorithm in that situation. The primary reason is that when the delay is small, the effect of augmented state-space on training is more severe than the model error introduced by delay-unawareness.

\begin{figure*}[t]
  \centering
  \includegraphics[width=0.9\linewidth]{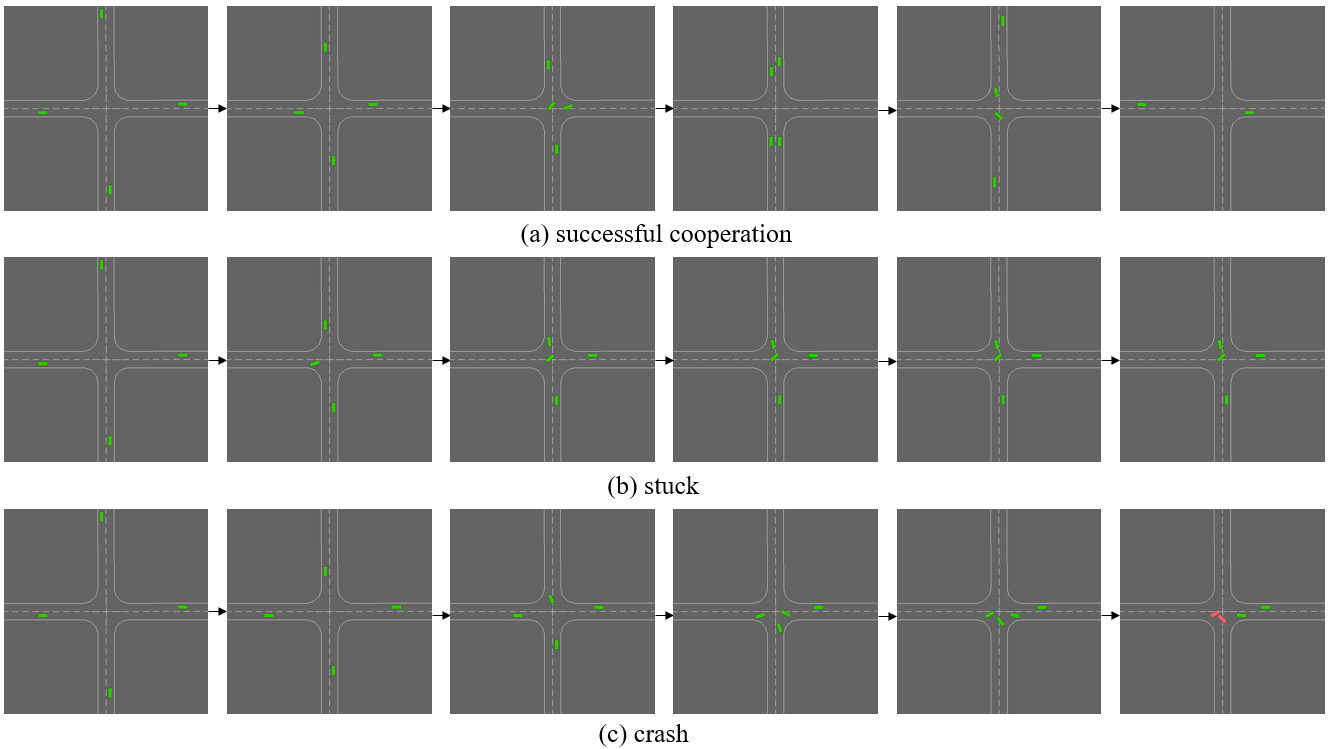}
  \caption{Typical outcomes of the unsignalized intersection scenario. In (a), the vehicles cooperated to pass the intersection by pairs since vehicles from opposite directions have non-intersecting routes; while in (b) and (c), vehicles failed to pass the intersection safely within the time limit.}
  \label{fig:outcomes}
\end{figure*}

\subsection{Traffic Environment}
\begin{figure}[h]
  \centering
  \includegraphics[width=0.5\linewidth]{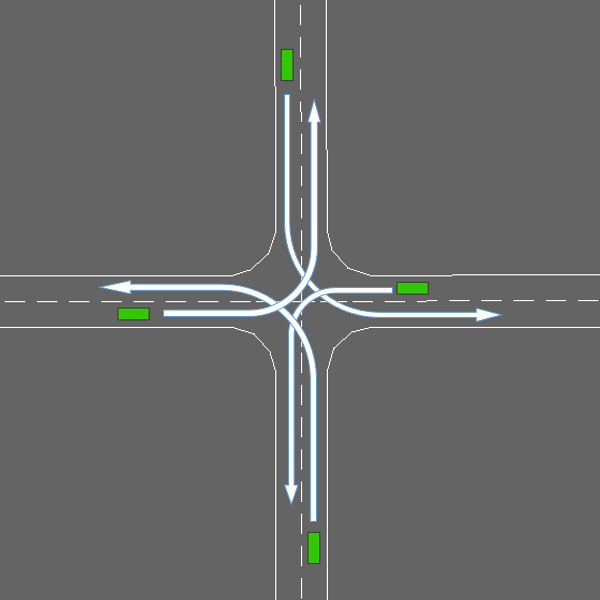}
  \caption{Unsignalized intersection scenario. The goal of the vehicles is to take a left turn at the intersection.}
  \label{fig:unsig}
\end{figure}

To show the practical value of delay-awareness, we construct an unsignalized intersection scenario that requires coordination of autonomous vehicles as shown in Fig.~\ref{fig:unsig}. This scenario consists of four vehicles coming from four directions (north, south, west, east) of the intersection, respectively. The goal of the vehicles is to take a left turn at the intersection. Vehicles are spawned at a random distance ($d^i \sim \mathcal{N}(\mu = 50 \text{ m}, \sigma = 10 \text{ m})$) from the intersection center with an initial velocity ($v^i_0 = 10$ m/s). Vehicles can observe the position and velocity of other vehicles as well as themselves. The routes are pre-defined and the vehicles can decide their longitudinal acceleration based on their policies.
The intersection is unsignalized so the vehicles need to coordinate to decide the sequence of passing. Vehicles are positively rewarded if all of them successfully finish the left turn and penalized if any collision happens. 

\begin{figure*}[t]
  \centering
  \includegraphics[width=0.9\linewidth]{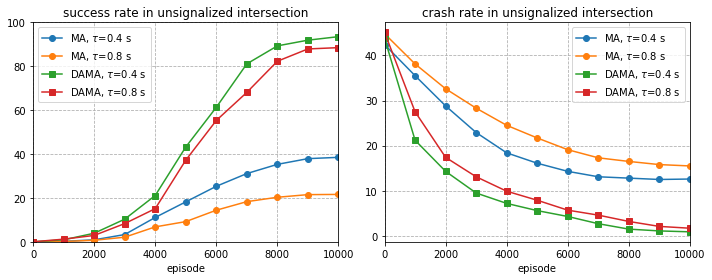}
  \caption{Success rate and crash rate of MA-DDPG and DAMA-DDPG in unsignalized intersection scenario. It is shown that delay-awareness drastically improve the performance of vehicles, both in success rate and crash rate. Delay-unaware agents suffer from huge model error and are not able to learn a good policy.}
  \label{fig:int_result}
\end{figure*}

In the traffic scenarios, the delay of an vehicle mainly includes: communication delay $\tau_1$, sensor delay $\tau_2$, time for decision making $\tau_3$ , actuator delay $\tau_4$. Since it has been proved that observation and action delays form the same mathematical problem~\cite{katsikopoulos2003markov}, we approximately assume that the total delay time is the sum of all the components: $\tau = \tau_1 + \tau_2 + \tau_3 + \tau_4$. Literature showed that the communication delay $\tau_1$ of vehicle-to-vehicle (V2V) systems with dedicated short-range communication (DSRC) devices can be minimal with a mean value of 1.1 ms~\cite{biswas2006vehicle, ammoun2006real} under good condition; the delay of sensors (cameras, LIDARs, radars, GPS, etc) $\tau_2$ is usually between 0.1 and 0.3 seconds~\cite{wang2018delay}; the time for decision making $\tau_3$ depends on the complexity of the algorithm and is usually minimal; the actuator delay $\tau_4$ for vehicle powertrain system and hydraulic brake system is usually between 0.3 and 0.6 seconds \cite{bayan2009brake, rajamani2011vehicle}. Adding them together, a conservative estimation of the total delay time $\tau$ of a vehicle would be roughly between 0.4 and 0.8 seconds, without communication loss. Thus, we test the delay-aware (DAMA-DDPG) and delay-unaware (MA-DDPG) algorithms under the delay $\tau$ of 0.4 and 0.8 seconds. 

There are three possible outcomes for each experiment: \textit{success}, \textit{stuck}, \textit{crash}, as shown in Fig~\ref{fig:outcomes}. We evaluate the performance of the learned policies based on the success rate and the crash rate. The results are shown in~\ref{fig:int_result}. It is shown that delay-awareness drastically improves the performance of vehicles, both in success rate and crash rate. The delay-aware agents successfully learn how to cooperate and finish both tasks without crash, while the delay-unaware agents suffer from huge model error introduced by delay and are not able to learn good policies: under 0.8 seconds delay, the success rate is less than 40\% for the unsignalized intersection task. The results are reasonable: consider a velocity of 10 m/s, the 0.8 seconds delay could cause a position error of 8 m, 
which injects huge uncertainty and bias to the state-understanding of the agents.
With highly-biased observations, the agents are not able to learn good policies to finish the task.


\section{Conclusion}
In this work, we propose a novel framework to deal with delays as well as non-stationary training issue of multi-agent tasks with model-free deep reinforcement learning. We formally define a general model for multi-agent delayed systems, Delay-Aware Markov Game, by augmenting standard Markov Game with agent delays while maintaining the Markov property. The solidity of this new structure is proved with the Markov reward process. For the agent training part, we propose a delay-aware algorithm that adopts the paradigm of centralized training with decentralized execution, and refer to it as delay-aware multi-agent reinforcement learning. Experiments are conducted in the multi-agent particle environment as well as a practical traffic simulator with autonomous vehicles. Results show that the proposed delay-aware multi-agent reinforcement learning algorithm greatly alleviate the performance degradation introduced by delay.

Though the delay problem in multi-agent systems is elegantly formalized and solved, the state augmenting procedure will increase the dimension of the problem. A promising direction for future work is to improve the sample efficiency by incorporating opponent modeling into the framework.

\appendix

\begin{restatable*}{theorem}{TheoremRttb2}
\label{the:equn_a}
A set of policy $\augm \pi: \mathcal{A} \times \pmb{\mathcal{X}} \to [0,1]$ interacting with $D\!A\!M\!G(E, \augm k)$ in the delay-free manner produces the same Markov Reward Process as $\augm \pi$ interacting with $M\!G(E)$ with $\augm k$ action delays for agents, i.e.
\begin{equation} \label{DMDP_equality_a}
    D\!A\!M\!R\!P(M\!G(E), \augm \pi, \augm n) = M\!R\!P(D\!A\!M\!G(E, \augm n), \augm \pi).
\end{equation}
\end{restatable*}

\begin{proof} For any $M\!G(E)= (\mathcal{S}, \mathcal{A}, \mathcal{O}, \rho, p,  r)$, we need to prove that the above two MRPs are the same. Referring to Def.~\ref{def:DAMG}~and~\ref{def:MRP}, for $M\!R\!P(D\!A\!M\!G(E, \augm k), \augm \pi)$, we have \\
(1) augmented state space \hspace{0.3cm} $\pmb{\mathcal{X}} = \mathcal{S} \times \mathcal{A}_{1}^{k_1} \times \dots \times \mathcal{A}_{N}^{k_N}$, \\
(2) initial distribution 
\begin{align*}
\augm \rho(\x_0) &= \augm \rho({s_0, a^1_0, \dots, a^1_{k_1-1}, \dots, a^N_0, \dots, a^N_{k_N-1}}) \\
&= \rho(s_0) \ \prod_{i=1}^{N}\prod_{j=0}^{k_i-1}\delta(a^i_j - c^i_j),
\end{align*}
(3)  transition kernel
\begin{align*}
&\augm \kappa(\x_{t+1} | \x_t) \\
= &\int_\mathcal{A} \pmb{p}(\x_{t+1} | \x_t, \augm a_t) \augm \pi(\augm a_t | \x_t) \ d \augm a_t \\
=&\int_\mathcal{A} p(s_{t+1} | s_t, a^{1, (t)}_t, \dots, a^{N, (t)}_t) \prod_{i=1}^{N}\prod_{j=1}^{k_i-1}\delta(a^{i, (t+1)}_{t+j} - a^{i, (t)}_{t+j}) \\
&\prod_{i=1}^{N} \delta(a^{i, (k+1)}_{t+k_i} - \augm a_t^i) \augm\pi(\augm a_t | \augm x_t) \ d \augm a_t  \\
=&p(s_{t+1} | s_t, a^{1, (t)}_t, \dots, a^{N, (t)}_t)  \\ 
&\prod_{i=1}^{N}\prod_{j=1}^{k_i-1}\delta(a^{i, (t+1)}_{t+j} - a^{i, (t)}_{t+j}) \prod_{i=1}^{N} \augm \pi_i(a^{i, (t+1)}_{t+k_i} | o_{t}^i),
\end{align*}
(4) state-reward function
\begin{align*}
\pmb{\bar r}_i(\x_t) &= \int_{\mathcal{A}_i} \augm r_i(\x_t, \augm a_t) \augm \pi_i(\augm a_t^i | \x_t) \ d \augm a_t^i \\
&=\int_{\mathcal{A}_i} r_i(s_t, a^i_t)\ \augm \pi_i(\augm a_t^i | \x_t) \ d \augm a_t^i \\
&= r_i(s_t, a^i_t)\ \int_{\mathcal{A}_i}\augm \pi_i(\augm a_t^i | \x_t) \ d \augm a_t^i \\
&=r_i(s_t, a^i_t),
\end{align*}
 for each agent $i$. Since the expanded terms of $M\!R\!P(D\!M\!G(E, n), \augm \pi)$ match the corresponding terms of $D\!A\!M\!R\!P(M\!G(E), \augm \pi, n)$ (Def.~\ref{def:DAMRP}), Eq.~\ref{DMDP_equality} holds.
\end{proof}

%





\ifCLASSOPTIONcaptionsoff
  \newpage
\fi


\bibliographystyle{IEEEtran}
\bibliography{main}

\begin{thebibliography}{10}
\providecommand{\url}[1]{#1}
\csname url@samestyle\endcsname
\providecommand{\newblock}{\relax}
\providecommand{\bibinfo}[2]{#2}
\providecommand{\BIBentrySTDinterwordspacing}{\spaceskip=0pt\relax}
\providecommand{\BIBentryALTinterwordstretchfactor}{4}
\providecommand{\BIBentryALTinterwordspacing}{\spaceskip=\fontdimen2\font plus
\BIBentryALTinterwordstretchfactor\fontdimen3\font minus
  \fontdimen4\font\relax}
\providecommand{\BIBforeignlanguage}[2]{{%
\expandafter\ifx\csname l@#1\endcsname\relax
\typeout{** WARNING: IEEEtran.bst: No hyphenation pattern has been}%
\typeout{** loaded for the language `#1'. Using the pattern for}%
\typeout{** the default language instead.}%
\else
\language=\csname l@#1\endcsname
\fi
#2}}
\providecommand{\BIBdecl}{\relax}
\BIBdecl

\bibitem{mnih2013playing}
V.~Mnih, K.~Kavukcuoglu, D.~Silver, A.~Graves, I.~Antonoglou, D.~Wierstra, and
  M.~Riedmiller, ``Playing atari with deep reinforcement learning,''
  \emph{arXiv preprint arXiv:1312.5602}, 2013.

\bibitem{silver2016mastering}
D.~Silver, A.~Huang, C.~J. Maddison, A.~Guez, L.~Sifre, G.~Van Den~Driessche,
  J.~Schrittwieser, I.~Antonoglou, V.~Panneershelvam, M.~Lanctot \emph{et~al.},
  ``Mastering the game of go with deep neural networks and tree search,''
  \emph{nature}, vol. 529, no. 7587, p. 484, 2016.

\bibitem{matignon2012coordinated}
L.~Matignon, L.~Jeanpierre, and A.-I. Mouaddib, ``Coordinated multi-robot
  exploration under communication constraints using decentralized markov
  decision processes,'' in \emph{Twenty-sixth AAAI conference on artificial
  intelligence}, 2012.

\bibitem{foerster2016learning}
J.~Foerster, I.~A. Assael, N.~De~Freitas, and S.~Whiteson, ``Learning to
  communicate with deep multi-agent reinforcement learning,'' in \emph{Advances
  in neural information processing systems}, 2016, pp. 2137--2145.

\bibitem{mordatch2018emergence}
I.~Mordatch and P.~Abbeel, ``Emergence of grounded compositional language in
  multi-agent populations,'' in \emph{Thirty-Second AAAI Conference on
  Artificial Intelligence}, 2018.

\bibitem{sukhbaatar2016learning}
S.~Sukhbaatar, R.~Fergus \emph{et~al.}, ``Learning multiagent communication
  with backpropagation,'' in \emph{Advances in neural information processing
  systems}, 2016, pp. 2244--2252.

\bibitem{peng2017multiagent}
P.~Peng, Q.~Yuan, Y.~Wen, Y.~Yang, Z.~Tang, H.~Long, and J.~Wang, ``Multiagent
  bidirectionally-coordinated nets for learning to play starcraft combat
  games,'' \emph{arXiv preprint arXiv:1703.10069}, vol.~2, 2017.

\bibitem{hernandez2017survey}
P.~Hernandez-Leal, M.~Kaisers, T.~Baarslag, and E.~M. de~Cote, ``A survey of
  learning in multiagent environments: Dealing with non-stationarity,''
  \emph{arXiv preprint arXiv:1707.09183}, 2017.

\bibitem{bu2008comprehensive}
L.~Bu, R.~Babu, B.~De~Schutter \emph{et~al.}, ``A comprehensive survey of
  multiagent reinforcement learning,'' \emph{IEEE Transactions on Systems, Man,
  and Cybernetics, Part C (Applications and Reviews)}, vol.~38, no.~2, pp.
  156--172, 2008.

\bibitem{agogino2004unifying}
A.~K. Agogino and K.~Tumer, ``Unifying temporal and structural credit
  assignment problems,'' in \emph{Proceedings of the Third International Joint
  Conference on Autonomous Agents and Multiagent Systems-Volume 2}.\hskip 1em
  plus 0.5em minus 0.4em\relax IEEE Computer Society, 2004, pp. 980--987.

\bibitem{matignon2012independent}
L.~Matignon, G.~J. Laurent, and N.~Le~Fort-Piat, ``Independent reinforcement
  learners in cooperative markov games: a survey regarding coordination
  problems,'' \emph{The Knowledge Engineering Review}, vol.~27, no.~1, pp.
  1--31, 2012.

\bibitem{hernandez2019survey}
P.~Hernandez-Leal, B.~Kartal, and M.~E. Taylor, ``A survey and critique of
  multiagent deep reinforcement learning,'' \emph{Autonomous Agents and
  Multi-Agent Systems}, vol.~33, no.~6, pp. 750--797, 2019.

\bibitem{lazarevic2006finite}
M.~Lazarevi{\'c}, ``Finite time stability analysis of pd$\alpha$ fractional
  control of robotic time-delay systems,'' \emph{Mechanics research
  communications}, vol.~33, no.~2, pp. 269--279, 2006.

\bibitem{hannah2018unbounded}
R.~Hannah and W.~Yin, ``On unbounded delays in asynchronous parallel
  fixed-point algorithms,'' \emph{Journal of Scientific Computing}, vol.~76,
  no.~1, pp. 299--326, 2018.

\bibitem{gu2003survey}
K.~Gu and S.-I. Niculescu, ``Survey on recent results in the stability and
  control of time-delay systems,'' \emph{Journal of dynamic systems,
  measurement, and control}, vol. 125, no.~2, pp. 158--165, 2003.

\bibitem{gong2016constrained}
S.~Gong, J.~Shen, and L.~Du, ``Constrained optimization and distributed
  computation based car following control of a connected and autonomous vehicle
  platoon,'' \emph{Transportation Research Part B: Methodological}, vol.~94,
  pp. 314--334, 2016.

\bibitem{bayan2009brake}
F.~P. Bayan, A.~D. Cornetto, A.~Dunn, and E.~Sauer, ``Brake timing measurements
  for a tractor-semitrailer under emergency braking,'' \emph{SAE International
  Journal of Commercial Vehicles}, vol.~2, no. 2009-01-2918, pp. 245--255,
  2009.

\bibitem{rajamani2011vehicle}
R.~Rajamani, \emph{Vehicle dynamics and control}.\hskip 1em plus 0.5em minus
  0.4em\relax Springer Science \& Business Media, 2011.

\bibitem{wang2018delay}
M.~Wang, S.~P. Hoogendoorn, W.~Daamen, B.~van Arem, B.~Shyrokau, and R.~Happee,
  ``Delay-compensating strategy to enhance string stability of adaptive cruise
  controlled vehicles,'' \emph{Transportmetrica B: Transport Dynamics}, vol.~6,
  no.~3, pp. 211--229, 2018.

\bibitem{artstein1982linear}
Z.~Artstein, ``Linear systems with delayed controls: A reduction,'' \emph{IEEE
  Transactions on Automatic control}, vol.~27, no.~4, pp. 869--879, 1982.

\bibitem{moulay2008finite}
E.~Moulay, M.~Dambrine, N.~Yeganefar, and W.~Perruquetti, ``Finite-time
  stability and stabilization of time-delay systems,'' \emph{Systems \& Control
  Letters}, vol.~57, no.~7, pp. 561--566, 2008.

\bibitem{astrom1994new}
K.~J. Astrom, C.~C. Hang, and B.~Lim, ``A new smith predictor for controlling a
  process with an integrator and long dead-time,'' \emph{IEEE transactions on
  Automatic Control}, vol.~39, no.~2, pp. 343--345, 1994.

\bibitem{matausek1999modified}
M.~R. Matausek and A.~Micic, ``On the modified smith predictor for controlling
  a process with an integrator and long dead-time,'' \emph{IEEE Transactions on
  Automatic Control}, vol.~44, no.~8, pp. 1603--1606, 1999.

\bibitem{mirkin2000extraction}
L.~Mirkin, ``On the extraction of dead-time controllers from delay-free
  parametrizations,'' \emph{IFAC Proceedings Volumes}, vol.~33, no.~23, pp.
  169--174, 2000.

\bibitem{niculescu2001delay}
S.-I. Niculescu, \emph{Delay effects on stability: a robust control
  approach}.\hskip 1em plus 0.5em minus 0.4em\relax Springer Science \&
  Business Media, 2001, vol. 269.

\bibitem{sutton2018reinforcement}
R.~S. Sutton and A.~G. Barto, \emph{Reinforcement learning: An
  introduction}.\hskip 1em plus 0.5em minus 0.4em\relax MIT press, 2018.

\bibitem{singh1994learning}
S.~P. Singh, T.~Jaakkola, and M.~I. Jordan, ``Learning without state-estimation
  in partially observable markovian decision processes,'' in \emph{Machine
  Learning Proceedings 1994}.\hskip 1em plus 0.5em minus 0.4em\relax Elsevier,
  1994, pp. 284--292.

\bibitem{walsh2009learning}
T.~J. Walsh, A.~Nouri, L.~Li, and M.~L. Littman, ``Learning and planning in
  environments with delayed feedback,'' \emph{Autonomous Agents and Multi-Agent
  Systems}, vol.~18, no.~1, p.~83, 2009.

\bibitem{firoiu2018human}
V.~Firoiu, T.~Ju, and J.~Tenenbaum, ``At human speed: Deep reinforcement
  learning with action delay,'' \emph{arXiv preprint arXiv:1810.07286}, 2018.

\bibitem{ramstedt2019real}
S.~Ramstedt and C.~Pal, ``Real-time reinforcement learning,'' in \emph{Advances
  in Neural Information Processing Systems}, 2019, pp. 3067--3076.

\bibitem{chen2020delayaware}
B.~Chen, M.~Xu, L.~Li, and D.~Zhao, ``Delay-aware model-based reinforcement
  learning for continuous control,'' 2020.

\bibitem{katsikopoulos2003markov}
K.~V. Katsikopoulos and S.~E. Engelbrecht, ``Markov decision processes with
  delays and asynchronous cost collection,'' \emph{IEEE transactions on
  automatic control}, vol.~48, no.~4, pp. 568--574, 2003.

\bibitem{schuitema2010control}
E.~Schuitema, L.~Bu{\c{s}}oniu, R.~Babu{\v{s}}ka, and P.~Jonker, ``Control
  delay in reinforcement learning for real-time dynamic systems: a memoryless
  approach,'' in \emph{2010 IEEE/RSJ International Conference on Intelligent
  Robots and Systems}.\hskip 1em plus 0.5em minus 0.4em\relax IEEE, 2010, pp.
  3226--3231.

\bibitem{tan1993multi}
M.~Tan, ``Multi-agent reinforcement learning: Independent vs. cooperative
  agents,'' in \emph{Proceedings of the tenth international conference on
  machine learning}, 1993, pp. 330--337.

\bibitem{papoudakis2019dealing}
G.~Papoudakis, F.~Christianos, A.~Rahman, and S.~V. Albrecht, ``Dealing with
  non-stationarity in multi-agent deep reinforcement learning,'' \emph{arXiv
  preprint arXiv:1906.04737}, 2019.

\bibitem{lowe2017multi}
R.~Lowe, Y.~Wu, A.~Tamar, J.~Harb, O.~P. Abbeel, and I.~Mordatch, ``Multi-agent
  actor-critic for mixed cooperative-competitive environments,'' in
  \emph{Advances in neural information processing systems}, 2017, pp.
  6379--6390.

\bibitem{lillicrap2015continuous}
T.~P. Lillicrap, J.~J. Hunt, A.~Pritzel, N.~Heess, T.~Erez, Y.~Tassa,
  D.~Silver, and D.~Wierstra, ``Continuous control with deep reinforcement
  learning,'' \emph{arXiv preprint arXiv:1509.02971}, 2015.

\bibitem{mnih2016asynchronous}
V.~Mnih, A.~P. Badia, M.~Mirza, A.~Graves, T.~Lillicrap, T.~Harley, D.~Silver,
  and K.~Kavukcuoglu, ``Asynchronous methods for deep reinforcement learning,''
  in \emph{International conference on machine learning}, 2016, pp. 1928--1937.

\bibitem{haarnoja2018soft}
T.~Haarnoja, A.~Zhou, P.~Abbeel, and S.~Levine, ``Soft actor-critic: Off-policy
  maximum entropy deep reinforcement learning with a stochastic actor,''
  \emph{arXiv preprint arXiv:1801.01290}, 2018.

\bibitem{highway-env}
E.~Leurent, ``An environment for autonomous driving decision-making,''
  \url{https://github.com/eleurent/highway-env}, 2018.

\bibitem{jang2016categorical}
E.~Jang, S.~Gu, and B.~Poole, ``Categorical reparameterization with
  gumbel-softmax,'' \emph{arXiv preprint arXiv:1611.01144}, 2016.

\bibitem{biswas2006vehicle}
S.~Biswas, R.~Tatchikou, and F.~Dion, ``Vehicle-to-vehicle wireless
  communication protocols for enhancing highway traffic safety,'' \emph{IEEE
  communications magazine}, vol.~44, no.~1, pp. 74--82, 2006.

\bibitem{ammoun2006real}
S.~Ammoun, F.~Nashashibi, and C.~Laurgeau, ``Real-time crash avoidance system
  on crossroads based on 802.11 devices and gps receivers,'' in \emph{2006 IEEE
  Intelligent Transportation Systems Conference}.\hskip 1em plus 0.5em minus
  0.4em\relax IEEE, 2006, pp. 1023--1028.

\end{thebibliography}
\end{document}